\RequirePackage{fix-cm}
\documentclass[twocolumn]{svjour3}          % twocolumn
\smartqed  % flush right qed marks, e.g. at end of proof
\usepackage{graphicx}
\usepackage[tight,footnotesize]{subfigure}
\usepackage{amssymb}
\usepackage{amsmath}
\usepackage{url}
\begin{document}

\title{Image color transfer to evoke different emotions based on color combinations%\thanks{Grants or other notes
}


\author{Li He         \and
        Hairong Qi\and
        Russell Zaretzki
}


\institute{Li He, Hairong Qi, Russell Zaretzki\at
			  University of Tennessee, Knoxville\\
              \email{\{lhe4,hqi,rzaretzk\}@utk.edu}           %  \\
}

\date{Received: date / Accepted: date}

\maketitle

\begin{abstract}
In this paper, a color transfer framework to evoke different emotions for images based on color combinations is proposed. The purpose of this color transfer is to change the ``look and feel'' of images, i.e., evoking different emotions. Colors are confirmed as the most attractive factor in images. In addition, various studies in both art and science areas have concluded that other than single color, color combinations are necessary to evoke specific emotions. Therefore, we propose a novel framework to transfer color of images based on color combinations, using a predefined color emotion model. The contribution of this new framework is three-fold. First, users do not need to provide reference images as used in traditional color transfer algorithms. In most situations, users may not have enough aesthetic knowledge or path to choose desired reference images. Second, because of the usage of color combinations instead of single color for emotions, a new color transfer algorithm that does not require an image library is proposed. Third, again because of the usage of color combinations, artifacts that are normally seen in traditional frameworks using single color are avoided. We present encouraging results generated from this new framework and its potential in several possible applications including color transfer of photos and paintings.
\end{abstract}
\vspace{-5mm}
\section{Introduction}

Among the many possible image-processing options, artists and scientists are increasingly interested in extracting ``emotion'' that images can evoke as well as changing the emotion by altering its colors. The pioneer work by Reinhard et al. \cite{Reinhard01} made color transfer possible by providing a reference image. Later, various color transfer algorithms have been proposed. However, only until recently was the first emotion-related color transfer algorithm proposed by Yang and Peng \cite{Yang08}. Their work followed the traditional color transfer framework but added a single color scheme for emotions.

Color as an emotion messenger has attracted enormous interests from researchers in different disciplines \cite{Kobayashi92,Whelan94,Eisemann00,Ou04,Ou04b,Wang06,Csurka10}. One may delight in the beautiful red and golden-yellow leaves of autumn, and in the magnificent colors of a sunset. The relationships between color and emotion is referred to as \textit{color emotion} \cite{emotionweb}.

In our daily life colors are never seen in isolation, but always presented together with other colors. This is true when we look at our surroundings from the inside of a building to the entire cityscape \cite{Ou04b}. Therefore, it is inappropriate to apply single color scheme to identify the emotion evoked by color images. For instance, in Kobayashi \cite{Kobayashi92}'s book \textit{Color Image Scale}, ``red'' may have multiple meanings, such as rich, powerful, luxurious, dynamic and mellow, depending on what color it is combined with. Hence, color combinations are always preferred over single color to evoke specific emotions.

After Reinhard \cite{Reinhard01}'s ground breaking work, color transfer algorithms have been extensively studied  \cite{Chang04,Greenfield03,Tai05,Pitie05,Xiao06,Xiao09,Xiang09,Chiou10,Dong10,Pouli11}. Reinhard's method is simple and efficient, but suffers from two problems. First, it could produce unnaturally looking results in cases where the input and reference images have different color distributions. Second, as the algorithm is based on simple statistics (mean and deviation), it could produce results with low fidelity in both scene detail and color distribution \cite{Xiao09}.

Although there have been several methods proposed to solve these two problems, they still need one or multiple reference images, while in most situations users may not have enough aesthetic knowledge to choose appropriate reference images and finding a correct reference image may become a time consuming task.

In this paper, we present a new emotion-changing color transfer framework based on \textit{color combinations} that do not need reference images as used in traditional color transfer algorithms. Because of the usage of color combinations instead of single color scheme used in \cite{Yang08}, we also need to develop a new color transfer algorithm. The proposed emotion transfer framework allows users to select their desired emotion by providing keywords directly, e.g., ``warm'', ``romantic'' and ``cool''.

We adopt Eisemann \cite{Eisemann00}'s color emotion scheme called \textit{Pantone color scheme} that contains 27 emotions with each emotion containing 24 three color combinations. This scheme is based on the early work of word association studies of color and emotion and the color harmony theory developed by Itten \cite{Itten62}.

\begin{figure}[]
\centering
    \includegraphics[width = 1\columnwidth]{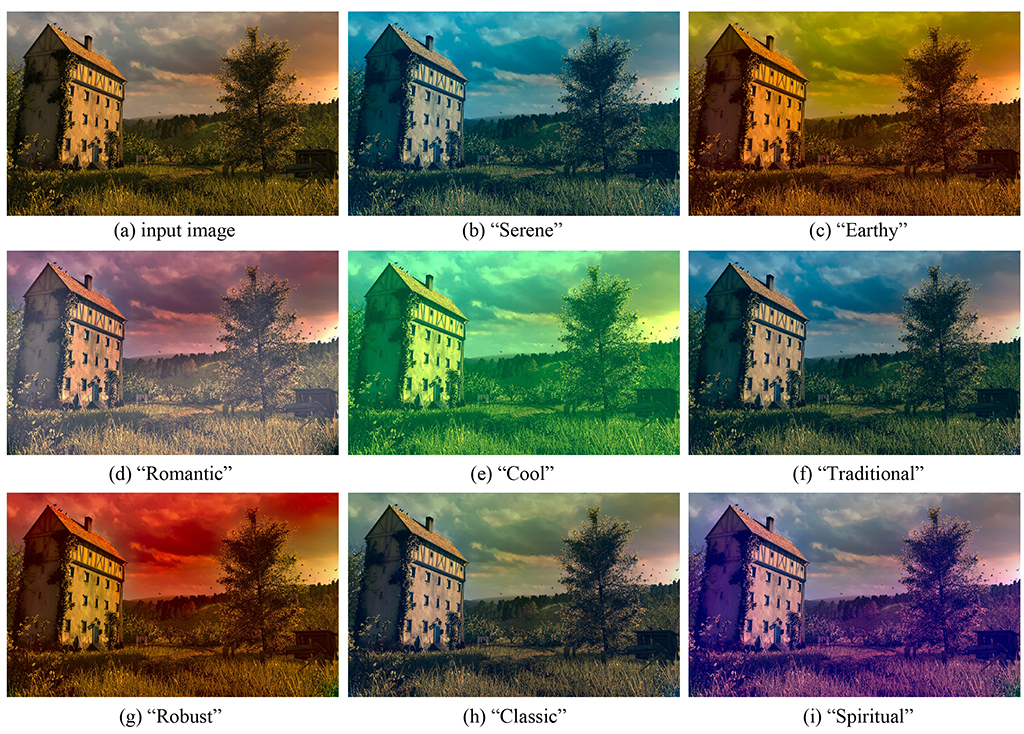}
\caption{Color transfer of an input image (a) to eight different themes (b)-(i) that evoke different emotions. The input image is an artwork called \textit{A lonely house} made by Michael Otto.}
\label{fig:teaser}
\vspace{-5mm}
\end{figure}

Due to the usage of color combinations, the proposed color transfer framework faces two challenging issues, which are, how to identify three main colors in the image and how to transfer main colors to destination color combinations. We resolve the first issue using an Expectation Maximization (EM) clustering algorithm. For the second issue, we model the color transfer process as two optimization problems, with one used to calculate target color combinations and the other
used to ensure the preservation of gradient of the input image, which is based on \cite{Xiao09}. Figure \ref{fig:teaser} shows an example result of proposed color transfer method.

The proposed color transfer framework could lead to many potential applications. For example, it can be used in field of art and design, such as photo / painting editing, transfer of the artist's emotion in painting to real photos, creative tone reproduction and color selection for industrial design. In addition, it can be used to color gray-scale images. Furthermore, for the purpose of color transfer with a reference image, the proposed color transfer algorithm may provide an output image which is better in capturing the emotion evoked in the reference image by involving a color emotion scheme.

The rest of the paper is organized as follows. Section 2 reviews previous work in related areas. Section 3 describes the proposed color combination-based emotion transfer framework in detail. Section 4 presents results and comparisons. Section 5 provides the discussion and future work.

\vspace{-6mm}

\section{Related Work}
\subsection{Color Emotion}
Recent models of color emotion include single color emotion and color combinations emotion. There have been two main research topics in the study of single color emotion: classification and quantification \cite{emotionweb}. Classification of single color emotion uses principal component analysis to reduce large number of colors to a small number of categories \cite{Kobayashi92,Sato00,Ou04}. The quantification of color emotion is studied firstly by~\cite{Sato00}, later by~\cite{Ou04} who provided a color emotion model with quantities on three color-emotion factors: activity, weight and heat. \cite{Ou04}'s study was then confirmed by~\cite{Wang06}. For color combinations, Kobayashi \cite{Kobayashi92} developed a color emotion model based on psychology studies. Ou et al. \cite{Ou04b} revealed a simple relationship of color and emotion that relies on single color emotion and color pair emotion. Lee et al. \cite{Lee07} used sets theory to evaluate color combinations. These early-stage studies on color emotion only consider color pairs.

Compared to single color emotion, color combinations provide a more appropriate and accurate way to describe color emotion in images. 
\vspace{-5mm}
\subsection{Emotion Semantics of Images}
Emotion semantics is the most abstract semantic structure in images, because it is closely related to cognitive models, culture background and aesthetic standards of users. Tanaka \cite{Tanaka00} concluded that the contribution of color, spatial frequency, and size to attractiveness follows the order of color $>$ size $>$ spatial frequency. Mao \cite{Mao03} proved that the fractal dimensions (FDs) of images are related to affective properties of an image. \cite{Wang06b} discovered that color has strong relationship with emotion word pairs. Therefore, we attempt to alter the emotion an image can stimulate by changing the colors. 
\vspace{-5mm}
% related color transfer
\subsection{Color Transfer}
We may classify existing color transfer methods into global and local algorithms, where ``global'' means the algorithm transfers colors using global statistic, e.g., global mean and global variance, while ``local'' means the algorithm transfers colors using different values for different regions of the input image.

The first global color transfer method was proposed by Reinhard et al. \cite{Reinhard01}. It shifts and scales the pixel values of the input image to match the mean and standard deviation of the reference image. This is done in the $l\alpha\beta$ opponent color space (CIELAB), which is an average decorrelated space that allows color transfer to take place independently in each channel \cite{Ruderman98}. Later, many global approaches are proposed using high-level statistical properties~\cite{Neumann05,Xiao09,Pitie05,Li10}.

A general problem with global color transfer approaches is that if the structure of the input image and the reference image are vastly different, the results could look unnatural. Reinhard et al. \cite{Reinhard01} proposed a local method based on the inverse distance weighting to remedy the problem. Chang et al. \cite{Chang04} proposed a color transfer method whereby colors are classified into categories derived through a psychophysical color naming study. Tai et al. \cite{Tai05} proposed a local transfer approach based on their soft color segmentation algorithm, where a modified EM algorithm was proposed. Chiou et al. \cite{Chiou10} proposed a local color transfer algorithm based on intrinsic component. Dong et al. \cite{Dong10} proposed a fast local color transfer algorithm with dominant color mapping based on Earth Mover's Distance (EMD). Huang and Chen \cite{Huang09} proposed a landmark-based sparse color representation for local color transfer.

The major difference between our approach and other color transfer methods not only resides in the color emotion model, but also that we do not need reference images. Therefore, existing color transfer algorithms may not satisfy the proposed emotion transfer task. The algorithm developed by Huang et al. \cite{Huang10} involved emotion elements, however, they only considered warm and cool aspects of color emotions, while emotion in images need more complex color emotion model. A method relatively close to ours is proposed by Yang and Peng \cite{Yang08} which provided an automatic mood-transferring framework for color images. They used single color emotion scheme described in \cite{Whelan94} to classify the input image to one of the 24 emotions. The major difference of our approach to this approach is that we use color combinations rather than single color for emotions. In addition, we do not need an image library for emotions. Because of these two differences, a new color transfer algorithm is developed. In addition to the above differences, we also select the best output image in a different way since we do not use reference images.

\vspace{-5mm}
\section{Color Transfer to Evoke Different Emotions}
\begin{figure}
\centering
    \includegraphics[width = 0.6\columnwidth]{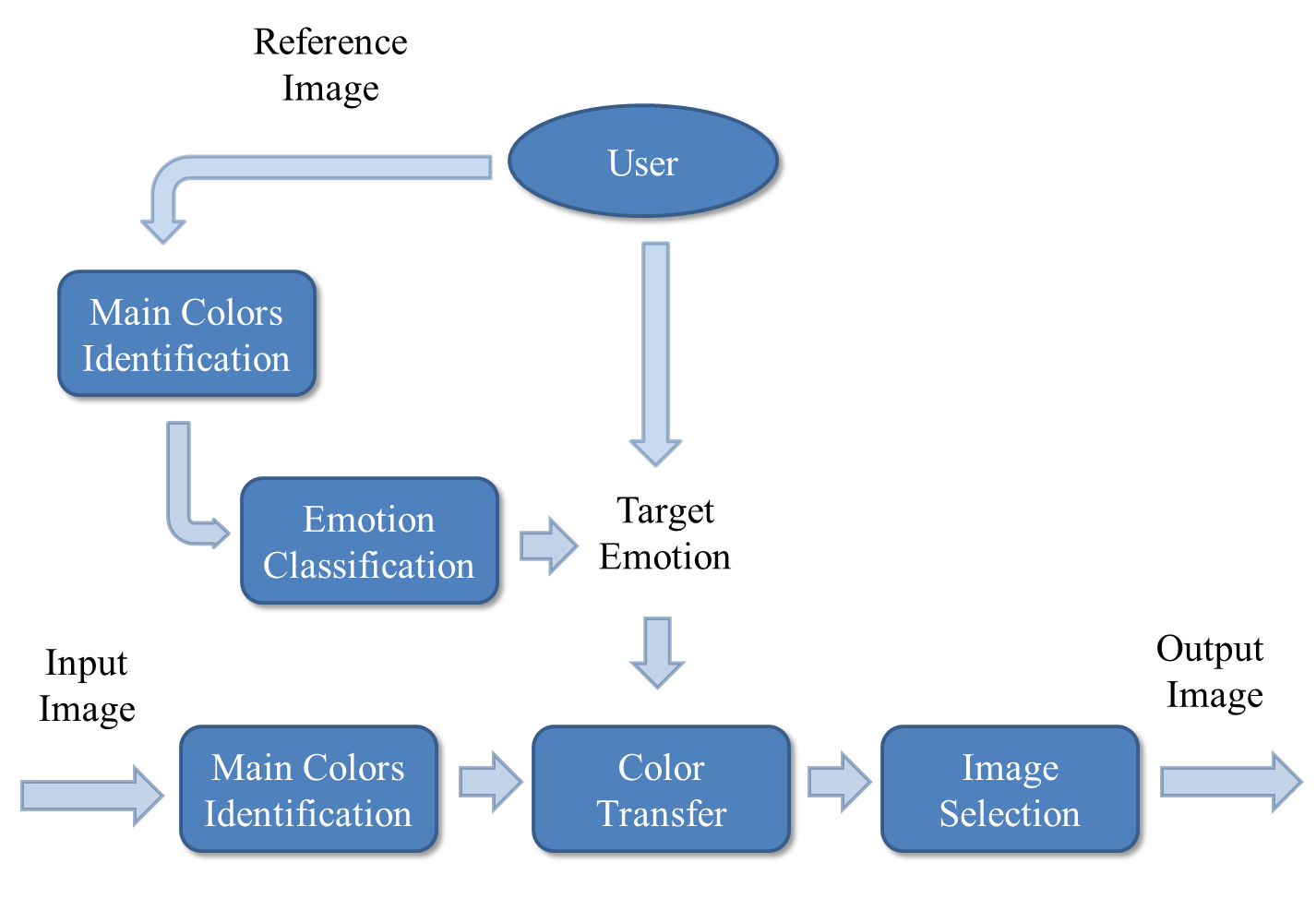}
\caption{Proposed color transfer process overview}
\label{fig:overview}
%\vspace{-6mm}
\end{figure}
Figure \ref{fig:overview} illustrates components involved in the proposed color transfer framework. Users can select target emotion by either providing a reference image or selecting an emotion keyword directly. If the user has a reference image, we can extract main colors (cluster centers in the color space) in reference image and use the closest scheme in color emotion scheme as target scheme. Either way, a specific scheme is selected and color combinations in this scheme are used for color transfer. Note that main colors mean the dominate, subordinate, and accent colors in the image.

Meanwhile, main colors in the input image are also extracted. Once we have main colors in the input image as well as target scheme, the color transfer algorithm transfers all colors in the input images to 24 output images (there are 24 three color combinations for each scheme). The final step selects the best output image out of the 24. Details of each step are described in the following sections.
\vspace{-5mm}
\subsection{Color Emotion Scheme}

\begin{figure}
\centering
    \includegraphics[width = 0.4\columnwidth]{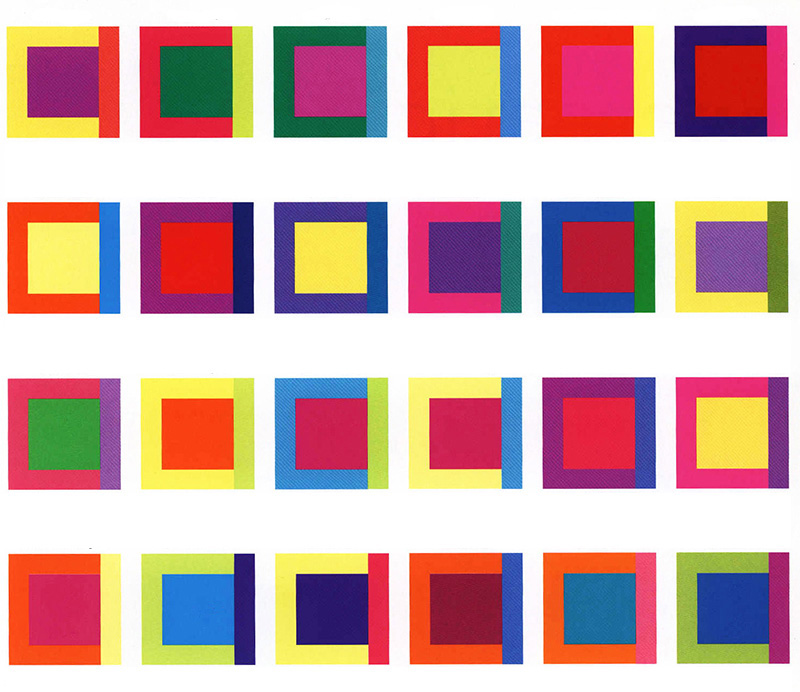}
\caption{24 sets of ``Playful'' color combinations \protect\cite{Eisemann00}}
\label{fig:playful}
%\vspace{-5mm}
\end{figure}

The \textit{Pantone color scheme} \cite{Eisemann00} contains 27 schemes with each scheme containing 24 three color combinations. In total, there are 648 three color combinations. The feeling that 27 schemes can evoke are Serene, Earthy, Mellow, Muted, Capricious, Spiritual, Romantic, Sensual, Powerful, Elegant, Robust, Delicate, Playful, Energetic, Traditional, Classic, Festive, Fanciful, Cool, Warm, Luscious\&Sweet, Spicy\&Tangy and Unique. Every color combination in each scheme includes three colors: dominant color, subordinate color and accent color. For example, the 24 color combinations of emotion ``Playful'' are illustrated in Figure \ref{fig:playful}. In each three color combination, the center color is the dominant color, the ``$\sqsubset$'' shape block is the subordinate color and the color shown in the right vertical bar is the accent color. All the colors in Pantone color scheme are in the CMYK space and we converted them to the CIELAB space.

%\vspace{-5mm}
\subsection{Main Colors Identification Using Clustering}
\begin{figure}[]
\centering
    \includegraphics[width = 1\columnwidth]{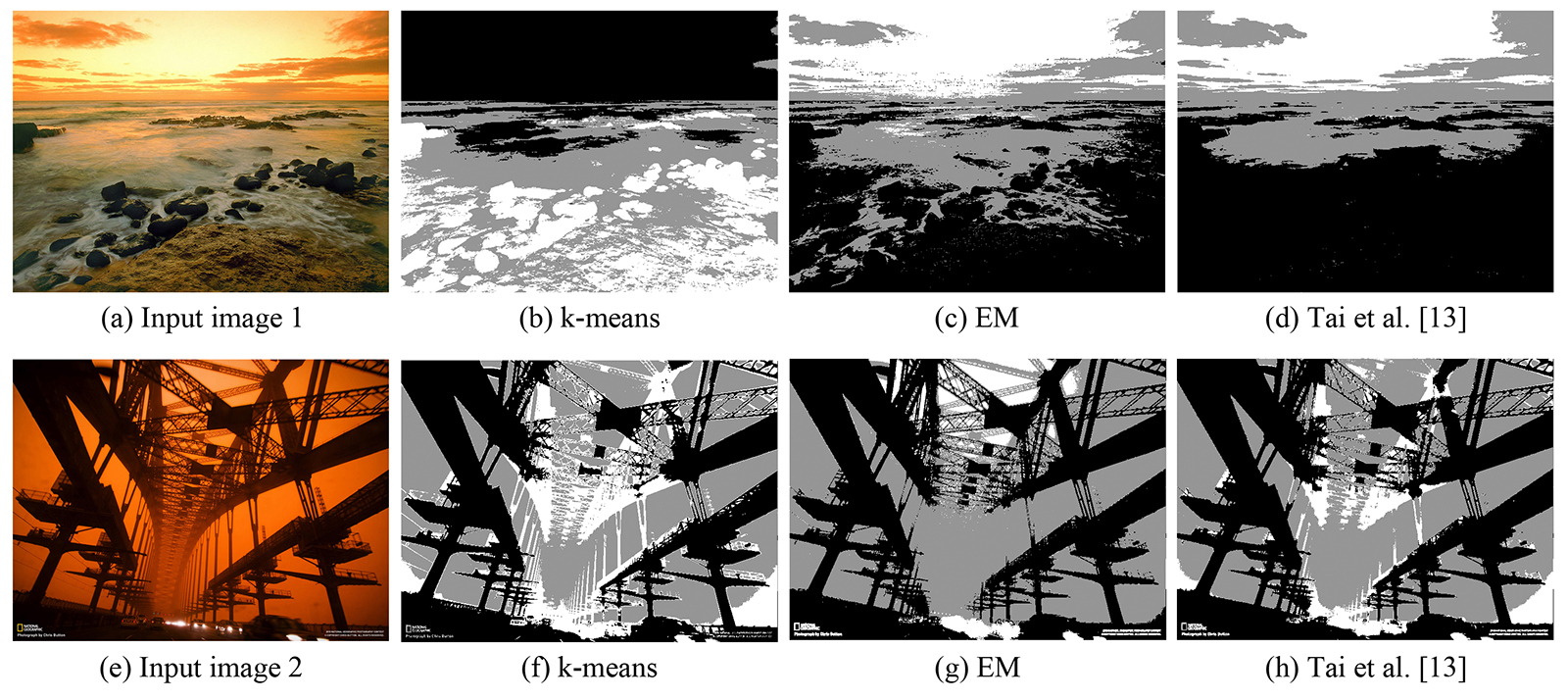}
\caption{Comparison of three different clustering algorithms. Black indicates the cluster of the dominant color, gray is the cluster of the subordinate color and white is the cluster of the accent color.}
\label{fig:cluster}
\vspace{-5mm}
\end{figure}
In order to match the three color combinations of the scheme, we need to extract the dominant, subordinate and accent colors in the input image. If the user chooses to use a reference image, the same process needs to be applied to extract the main colors of the reference image. We adopt the Expectation-Maximization (EM) algorithm in the CIELAB space.

We choose EM after comparing k-means, EM and improved EM \cite{Tai05}, as shown in Figure \ref{fig:cluster}\footnote{All input images in this paper are from National Geography.}. Compared to EM, a key limitation of k-means is its cluster model. The clusters are expected to be of similar size, so that the assignment to the nearest cluster center is the correct assignment. EM is more flexible by taking into consideration of both variances and covariances of clusters \cite{kmean}. In addition, in order to transfer colors of the input image, we want to segment the image better in color-wise, not in object-wise. Comparing to the result of EM (Figure \ref{fig:cluster}), the result of k-means segments the input image better in object-wise (it separates the sky and water precisely), while EM segments images better in color-wise. Because of this reason, the weights of Gaussian components in this algorithm can be naturally used as weights of the dominant, subordinate and accent colors.

We also implement the improved EM algorithm proposed by \cite{Tai05}, in which the spatial information is added. As shown in Figures \ref{fig:cluster}, Tai's algorithm has better region smoothness because of the spatial filter. However, in many images we might not want this feature. For instance, result of Tai's algorithm merged rocks with the ground and water surrounded into one region, while result of EM separated them.
\vspace{-5mm}
\subsection{Color Scheme Classification}
If the user provides a reference image, we classify the reference image to a specific related scheme first. After identifying the three main colors in the reference image, a Euclidean distance measure is used to classify the emotion:
\begin{equation}\label{eq:classify}
    \min_{i}(\sum_{j=1}^{24}{\sum_{k=1}^{3}{w_{k}\|\mathbf{C}^{k}_{R}-\mathbf{C}^{k}_{P_{ij}}\|_{2}^{2}}})
\end{equation}
where $i=1,2,\cdots,27$ is the $i^{th}$ scheme of the 27 schemes, $j=1,2,\cdots,24$ is the $j^{th}$ combination of the 24 color combinations in each scheme, $w_{k},k=1,2,3$ are weights of the $k^{th}$ cluster (Gaussian component) generated by the EM algorithm, $\mathbf{C}^{k}_{R},k=1,2,3$ are the three main colors in the reference image, $\mathbf{C}^{k}_{P_{ij}}$ is the $k^{th}$ color in the $j^{th}$ Pantone three color combination of the $i^{th}$ scheme. 

The scheme with the minimum distance is identified as the scheme of the reference image.

Now we have three clusters in the input image and a target emotion either specified by the user or identified in a reference image. Then, the emotion of the input image is transferred to the target emotion.

In this process, we use three guidelines to formulate the problem:
\begin{enumerate}
\item The transferred colors should still reside within the CIELAB space.
\item It is more important to guarantee the closeness between the transferred dominant cluster center to the dominant color in the color combination as compared to that of the subordinate or accent color.
\item It is well known that the human visual system is more sensitive to local intensity differences than to intensity itself \cite{Josa71}. Thus preserving the color gradient is necessary to scene fidelity \cite{Xiao09}.
\end{enumerate}

\begin{figure*}[]
\centering
    \includegraphics[width = 0.8\textwidth]{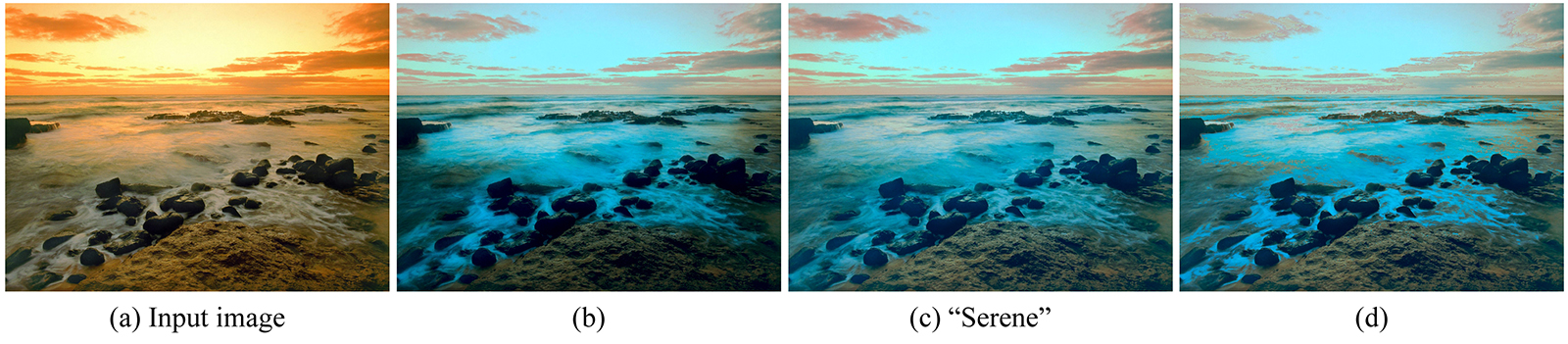}
\caption{Steps in transferring to ``Serene'' scheme. (b) transfer without limitation of cluster center movement (c) transfer with limit of cluster center movement (d) transfer without gradient preservation step.}
\label{fig:transfer}
%\vspace{-5mm}
\end{figure*}

\subsubsection{Calculation of Target Color Combinations}
\label{sec:cal}
The first step in Figure \ref{fig:toverview} is to calculate target color combinations. As we can see in Figure \ref{fig:transfer}, if we move the cluster centers of the input image to the exact Pantone color combination, the resulting image is darker than the input image. However, if we limit the movement of cluster centers, the resulting image is almost at the same brightness level compared to that of the input image.

We formulate the calculation of target color combinations as an optimization problem:
\begin{equation}\label{eq:centers}
\begin{aligned}
\min_{\mathbf{\delta}}\quad & f(\mathbf{\delta})=\sum_{k=1}^{3}{w_{k}(\|\mathbf{C}^{k}_{I}+\mathbf{\delta}^{k}-\mathbf{C}^{k}_{P_{ij}}\|_{2}^{2})}\\
\textrm{s.t.}\quad & l\alpha\beta_{\textrm{min}} \leq (\mathbf{I}^{k}_{\textrm{min}}+\mathbf{\delta}^{k}) \leq l\alpha\beta_{\textrm{max}}\\
\quad & l\alpha\beta_{\textrm{min}} \leq (\mathbf{I}^{k}_{\textrm{max}}+\mathbf{\delta}^{k}) \leq l\alpha\beta_{\textrm{max}}\\
\end{aligned}
\end{equation}
where $\mathbf{C}^{k}_{I},k=1,2,3$ are the three cluster centers of the input image calculated by the EM algorithm. $\mathbf{\delta}^{k}$ is the movement of each cluster center. $l\alpha\beta_{\textrm{min}}$ and $l\alpha\beta_{\textrm{max}}$ are the range of each dimension in the CIELAB space. $\mathbf{I}^{k}_{\textrm{min}}$ and $\mathbf{I}^{k}_{\textrm{max}}$ are the minimum and maximum values of the $k^{th}$ cluster in each dimension, respectively. $\mathbf{C}^{k}_{I^{'}},k=1,2,3$ are colors of the target color combination.

This optimization problem is designed to satisfy the guidelines (1) and (2). The condition in Eq. \ref{eq:centers} satisfies the first guideline, which guarantees all colors stay within range after color transfer. In addition, by minimizing the 2-norm distance in Eq.~\ref{eq:centers}, target color combinations are moved as close as possible to desired Pantone color combinations. Weights $w_{i}$ put different weights on the dominant, subordinate and accent colors which helps the minimization process to consider more about the movement of the dominant color.

The interior point algorithm described in \cite{Waltz06} is used to solve this optimization problem. 

Finally, target color combinations are calculated using Eq.~\ref{eq:ncenters}.
\begin{equation}\label{eq:ncenters}
    \mathbf{C}^{k}_{I^{'}}=\mathbf{C}^{k}_{I}+\mathbf{\delta}^{k}
\end{equation}

\subsubsection{Pixel Update}
\label{sec:update}
After calculation of target color combinations, the second step in Figure \ref{fig:toverview} updates all pixels in the input image:
\begin{equation}\label{eq:update}
    \mathbf{I}^{'k}_{xy}=\mathbf{I}^{k}_{xy}-\mathbf{C}^{k}_{I}+\mathbf{C}^{k}_{I^{'}}
\end{equation}
where $\mathbf{I}^{k}_{xy}$ and $\mathbf{I}^{'k}_{xy}$ are pixels of the $k^{th}$ clusters in the input and updated (intermediate) images, respectively.

%\begin{figure*}[]
%\centering
%    \includegraphics[width = 150mm]{fig8.jpg}
%\caption{Pixel update and gradient preservation: Red, green and blue are convex hulls of dominate, subordinate and accent clusters, respectively. Cluster centers of the input image are connected by yellow lines and target cluster centers are connected by magenta lines.}
%\label{fig:update}
%\end{figure*}

\subsubsection{Gradient Preservation}
The final step of color transfer is gradient preservation. Let us first observe the differences in the transfer result without (Figure \ref{fig:transfer}(d)) or with (Figure \ref{fig:transfer}(c)) gradient preservation as compared to the input image. We can see the artifacts in Figure \ref{fig:transfer}(d), such as the edge between cloud and sky and the edge between water and sky. However, the transfer result with gradient preservation has no artifacts in edges and has average brightness and contrast compared to that of the input image. As required by guideline (3), preserving the gradient is necessary to scene fidelity.

\begin{figure}[]
\centering
    \includegraphics[width = 1\columnwidth]{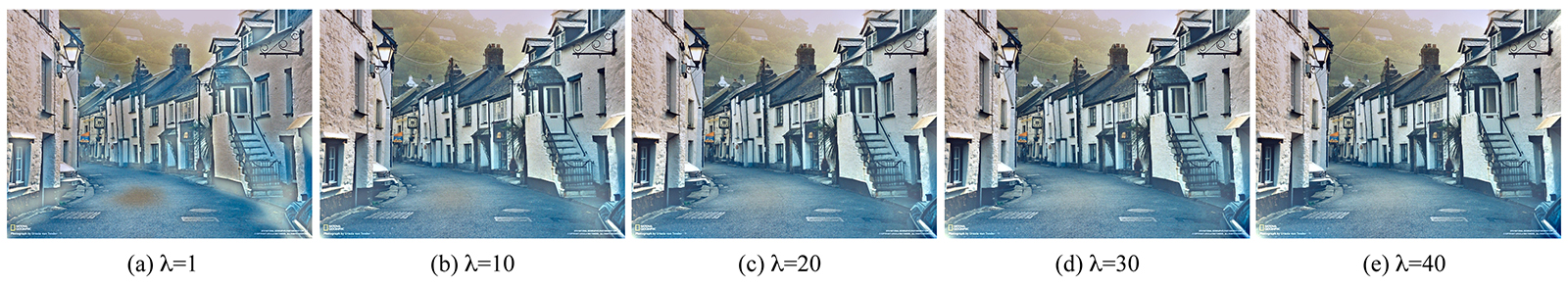}
\caption{Impact of $\lambda$ values on color transfer. The input image is transferred to "Muted" scheme.}
\label{fig:grad-example}
\end{figure}

\begin{figure}[]
\centering
    \includegraphics[width = 0.9\columnwidth]{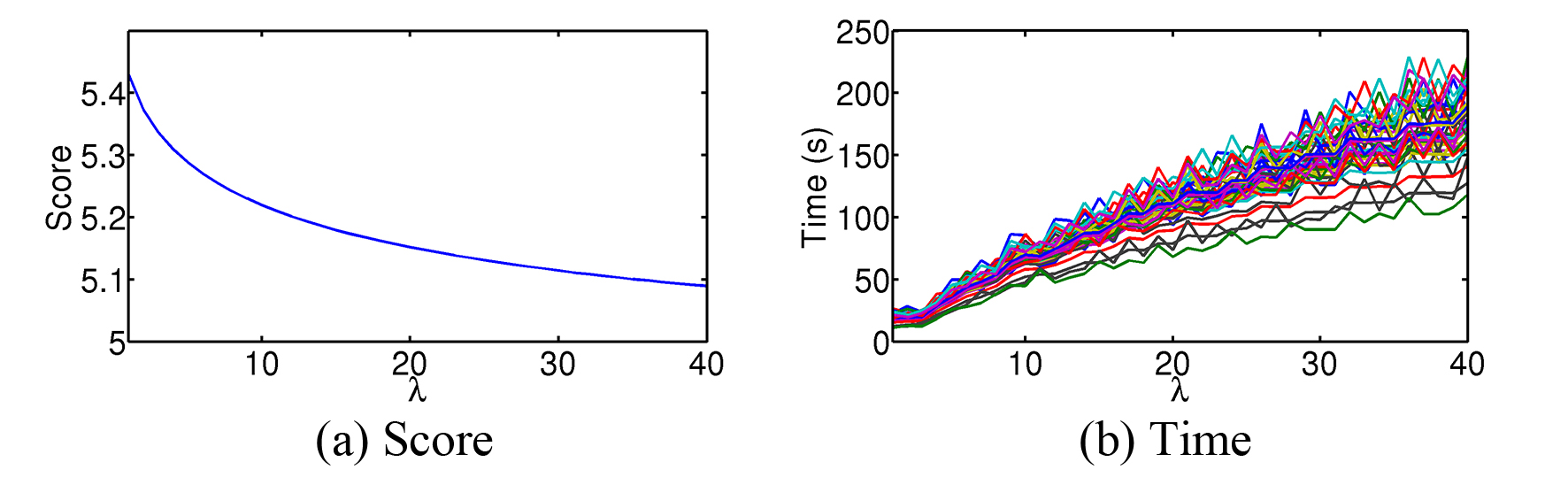}
\caption{Impact of $\lambda$ values on the color transfer scores and time. Each of 4 input images are transferred to 10 different schemes (color combinations). (a) illustrates average color transfer score, (b) illustrates color transfer time of 40 color transfers.}
\label{fig:grad}
%\vspace{-5mm}
\end{figure}

To preserve gradient, we use the algorithm proposed by \cite{Xiao09}
\small
\begin{equation}\label{eq:gradient}
\begin{aligned}
    \min_{\mathbf{O}_{xy}} &\sum_{x}\sum_{y}(\mathbf{O}_{xy}-\mathbf{I}^{'}_{xy})^{2}\\
&+\lambda\sum_{x}\sum_{y}[(\frac{\partial{\mathbf{O}_{xy}}}{\partial{x}}-\frac{\partial{\mathbf{I}_{xy}}}{\partial{x}})^{2}
+(\frac{\partial{\mathbf{O}_{xy}}}{\partial{y}}-\frac{\partial{\mathbf{I}_{xy}}}{\partial{y}})^{2}]
\end{aligned}
\end{equation}
\normalsize
where $\mathbf{I}_{xy}$, $\mathbf{I}^{'}_{xy}$, and $\mathbf{O}_{xy}$ are pixels in the input image, the intermediate image, and the output image, respectively. $x$ and $y$ are the horizontal and vertical axes of the image. $\lambda$ is a coefficient weighting the importance of gradient preservation and new colors.

The first term of Eq.~\ref{eq:gradient} ensures the output image is as similar as possible to the intermediate image. The second term of Eq.~\ref{eq:gradient} maintains the gradient of the output image as close as possible to the gradient of the input image. This optimization problem is solved by gradient descend method. 

\cite{Xiao09} set $\lambda$ equal to $1$ in their paper, however, the scene fidelity is not high enough in this application when $\lambda=1$. Impact of different $\lambda$ values on color transfer result is demonstrated in Figure \ref{fig:grad-example}. We also tested the impact of different $\lambda$ values on color transfer scores ($E(j)$ of Eq.~\ref{eq:output}) and color transfer time. In order to balance the color transfer scores and time, we choose $\lambda=20$ in this paper. In addition, throughout our experiment, we see consist trend of score on different image content (with different $\lambda$).  

\subsection{Output Image Selection}
\label{sec:output}
When a user selects a specific scheme, we transfer the input image to 24 output images based on the 24 color combinations of that scheme. Then the final output images is selected by evaluating content similarity of those images to the input image (in terms of luminance histogram) and distance of main colors (cluster centers) to the Pantone color combinations.

To measure the difference in luminance between the input and output images, we use the following \cite{Yang08}:
\small
\begin{equation}\label{eq:bright}
    d_{lumin}(\mathbf{I},\mathbf{O}_{j})=\frac{\sum_{x=1}^{width}\sum_{y=1}^{height}|l_{I_{x,y}}-l_{O_{(j)x,y}}|}{width\cdot height}
\end{equation}
\normalsize
where $l_{I_{x,y}}$ and $l_{O_{(j)x,y}}$ are the $l$ values (in $l\alpha\beta$ space) of the input and output images at pixel $(x,y)$, respectively. $\mathbf{I}$ is the input image and $\mathbf{O}_{j}$ is $j^{th}$ combination of the 24 color combinations. $width$ and $height$ represent the width and height of the input and output images.

To measure how close the cluster centers in the output image to the target Pantone color combinations, we use the following equation:
\small
\begin{equation}\label{eq:emotion}
    d_{color}(\mathbf{C}_{O},\mathbf{C}_{P_{ij}})=\sum_{k=1}^{3}|\mathbf{C}^{k}_{O}-\mathbf{C}^{k}_{P_{ij}}|
\end{equation}
\normalsize
where $\mathbf{C}^{k}_{O}$ are colors in the target color combination (cluster centers in the output images).

Finally, the best output image is selected by:
\small
\begin{equation}\label{eq:output}
\begin{aligned}
    \min_{j}
    E(j)=\gamma d_{lumin}(\mathbf{I},\mathbf{O}_{j}) + (1-\gamma)d_{color}(\mathbf{C}_{O_{j}},\mathbf{C}_{P_{ij}})
\end{aligned}
\end{equation}
\normalsize
where $j=1,2,...,24$, and $\gamma$ are weighting factors to combine two types of differences into a unified metric. In order to avoid unnatural looking of the final output image, we want to emphasize more the content similarity (measured by luminance difference). Therefore, we choose the $\gamma$ value of $0.7$ based on empirical study. In addition, the $\gamma$ value is not dependent on the image content with our test images. The image with minimum $E(j)$ value is chosen as the final output image. 
\vspace{-5mm}
\section{Results and Comparisons}
\begin{figure}[]
\centering
    \includegraphics[width = 1\columnwidth]{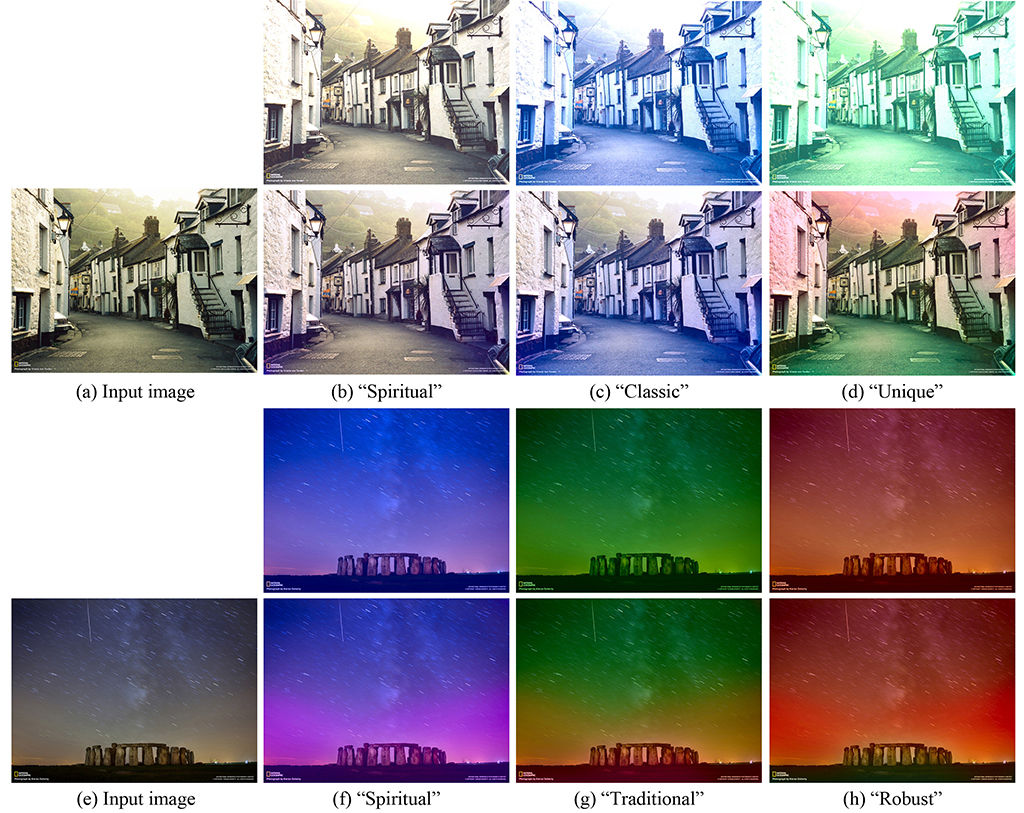}
    \vspace{-3mm}
\caption{Comparison of color transfer using color combination and single color. For 4 input images, each image is transferred to three schemes. }
\label{fig:result}
\vspace{-4mm}
\end{figure}

Our approach transfers the color of the input images using color combinations. Three applications of this approach are shown first in this section, including color transfer of an artwork, photos and a painting. Next, we show the effectiveness of the proposed method with a user study. This is followed by a comparison between color transfer method based on color combinations and the color transfer method based on single color is presented. At last, we compare our method with the traditional color transfer algorithms.

\textbf{Color transfer to evoke different emotion}. Figure \ref{fig:teaser} shows an example of the proposed color transfer of an artwork. The input image is transferred to eight target schemes, including Serene, Earthy, Romantic, Cool, Traditional, Robust, Classic and Spiritual. Figure \ref{fig:result} shows an application of our approach on photos.

\begin{figure}
\centering
    \includegraphics[width = 1\columnwidth]{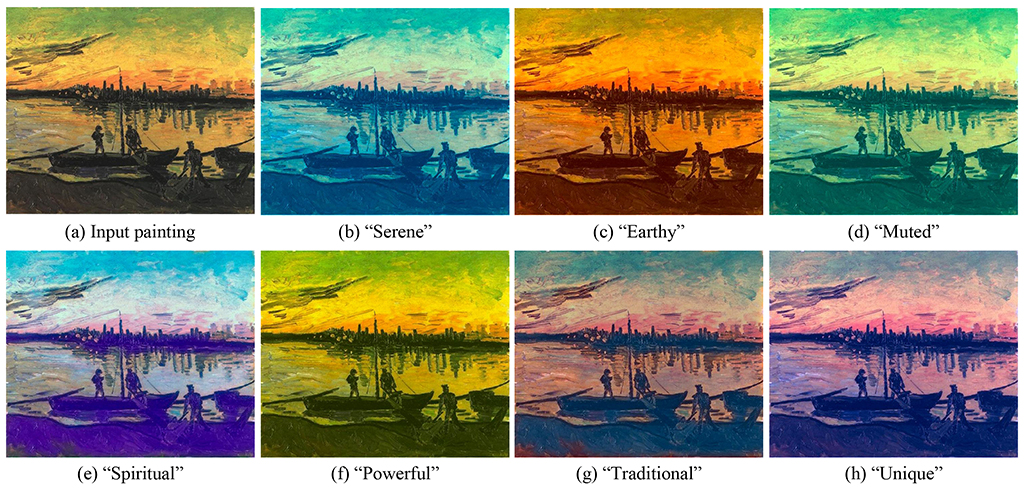}
\caption{Proposed color transfer of a painting. The painting (a) is Vincent van Gogh's \textit{Coal Barges}}
\label{fig:paint}
\vspace{-8mm}
\end{figure}

Figure \ref{fig:paint} shows another application of our approach on paintings, where a painting is transferred to seven alternative color schemes.

\textbf{User study}. In order to evaluate the effectiveness of our color transfer approach, a user study is designed to evaluate the results. 5 photos are transferred to 24 color themes using proposed approach. Each user is asked to choose the preference image of a given emotion/feeling in a pair of images, where the pair contains two randomly selected color transfer results of an input image. For example, for a given feeling ``Serene'', the user has to choose one image between two images in Figure \ref{fig:teaser} (b)(c), where Figure~\ref{fig:teaser} is randomly chosen for comparison. At the same time, a detail explanation of each feeling is displayed in the evaluation interface for user to further understand the meaning of the given feeling. For instance, the explanation of ``Serene'' is ``calm, peaceful, quiet, clean''. In total, 16 users participated in the study and each user evaluated 30 image pairs, totally 480 image pairs are evaluated. 

\begin{figure}
\centering
    \includegraphics[width = 0.8\columnwidth]{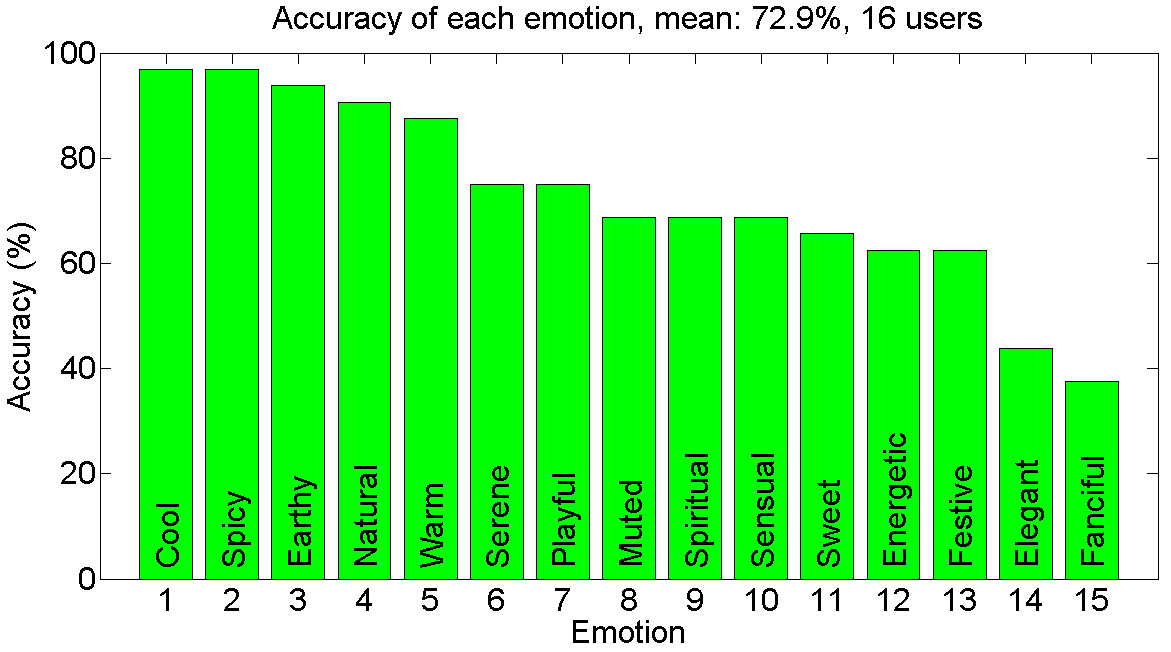}
\caption{User study - accuracy of each emotion.}
\label{fig:user-emotion}
\end{figure}

\begin{figure}
\centering
    \includegraphics[width = 0.8\columnwidth]{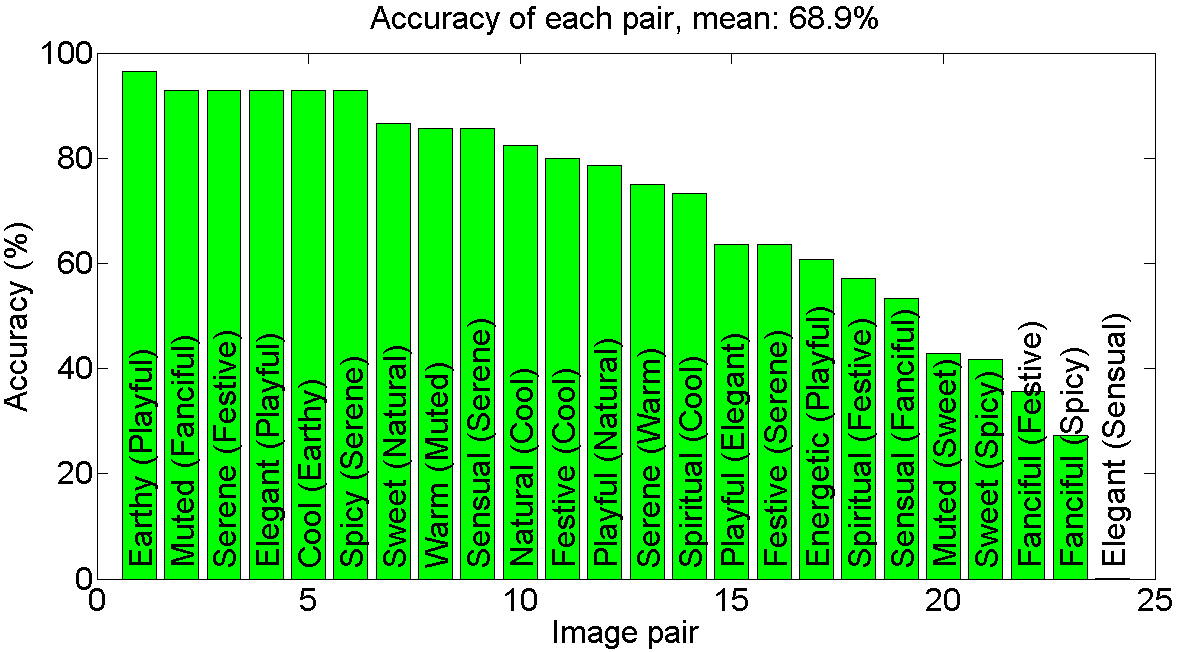}
\caption{User study - accuracy of each emotion pair.}
\label{fig:user-pair}
\vspace{-5mm}
\end{figure}

The evaluation results are shown in Figures~\ref{fig:user-emotion} and~\ref{fig:user-pair}. In Figure~\ref{fig:user-emotion}, We evaluated the accuracy rate of each emotion. 15 out of 24 emotions are shown in the figure, the rest 9 emotions are not shown because they were only selected in a few tests. We can see from the results, for easy-to-understand feelings such as ``cool'' and ``warm'' the accuracy is high. For ambiguous feelings such as ``Energetic'' and ``Fanciful'' the accuracy is low. Furthermore, we show the accuracy rate of each specific pair in Figure~\ref{fig:user-pair}. Pairs that contain strong contrast have high accuracy, such as ``Earthy-Playful'' and ``Serene-Festive''. Pairs that are ambiguous have low accuracy, such as ``Fanciful-Spicy'' and ``Sweet-Spicy''. Overall, we are able to achieve the average accuracy of $\sim70\%$ (compared to $50\%$ if selected randomly). We may improve the result in the future with a better description of each emotion. 

\textbf{Comparison with the single color method}. Figure \ref{fig:result} shows comparison of color transfer using color combinations and single colors. Color transfer using single color is implemented by moving the mean of $l, \alpha, \beta$ values of the input images to the dominate color of Pantone color combination. Compared to the transfer results using single colors, results using color combinations have several advantages. Firstly, using color combinations to represent emotion an image can evoke allows color transfer to be carried out separately in different regions of images, producing more colorful and emotionally rich images. As shown in Figures \ref{fig:result}(d), colors of the road, houses and trees are transferred to different colors separately using color combinations, while those objects are transferred to similar colors using single color. Similarly, as shown in Figures \ref{fig:result}(f), the colors of sky, lights near the ground and stones are transferred to different colors separately using color combinations, producing more colorful images compared to transfer results using single colors. Secondly, artifacts are eliminated using color combinations. For example, as shown in Figure \ref{fig:result}(d), colors of the wall and trees are transferred to an unnatural look (green) using single colors, while colors of the wall and trees still look natural using color combinations. Finally, using color combinations can avoid out of range problem. In addition, several transfer results are too bright using single colors, while transfer results using color combinations successfully avoid this problem.


\begin{figure*}[]
\centering
    \includegraphics[width = 0.8\textwidth]{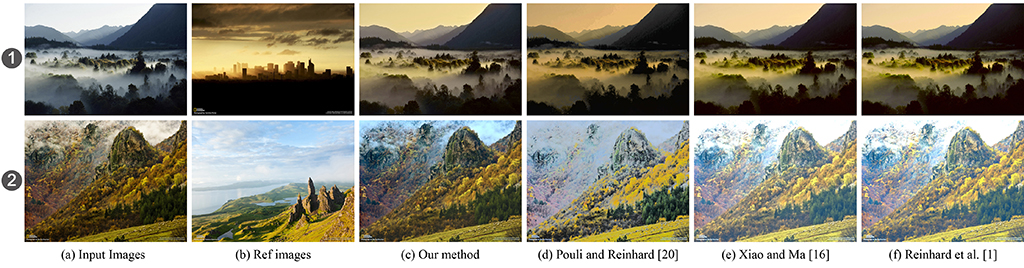}
\caption{Comparison between existing color transfer methods and the proposed color transfer method. }
\label{fig:ct}
\vspace{-5mm}
\end{figure*}

\textbf{Comparison with color transfer}. Although the purpose of our color transfer algorithm is different from traditional color transfer methods, we also compare our method (not including color emotion scheme) to three representative color transfer algorithms (\cite{Reinhard01,Xiao09,Pouli11}). If the user provides a reference image, the purpose of our algorithm is to transfer the ``emotion'' that image evoke to similar ``emotion'' of the reference image evokes, while the purpose of traditional color transfer methods is to transfer the ``colors'' in the input image to ``colors'' of the reference image. Figure \ref{fig:ct} shows the color transfer results, with two input images and two reference images. Compared to other methods, our method is able to blend the colors of the reference image, while preserving the look and details of the input image. For the first input image, Reinhard \cite{Reinhard01} and Xiao \cite{Xiao09}'s results suffer from high color saturation in the trees region compared to that of the input image. Pouli and Reinhard \cite{Pouli11}'s method suffers from artifacts in the sky and losing gradient of the input image in the bottom area. However, our method is able to avoid these problems. For the second input image, result images of Reinhard \cite{Reinhard01} and Xiao \cite{Xiao09}'s methods are too bright compared to the input image, resulting in losing details in highlight region. Pouli and Reinhard \cite{Pouli11}'s method suffers from losing gradient of the input image. Again, our method successfully avoids artifacts and maintain similar brightness level as compared to the input image.

\vspace{-3mm}
\section{Discussion and Future Work}
In this paper, we proposed a novel color transfer framework to evoke different emotions based on color combinations. Unlike traditional color transfer algorithms, users are not required to provide reference images. They can choose target emotion terms and the desired emotion is transferred automatically. This process is achieved by using a color scheme, in which each scheme is represented using three color combinations. 

Results and user study showed our approach is able to alter emotion evoked by photos and paintings, providing emotionally rich images for art and design purpose. 

While our color transfer method can produce images that convey rich emotions, there are still several limitations. For example, current weights of the dominant, subordinate and accent colors are decided automatically by the weights of Gaussian components. However, sometime the dominant color to human visual system may not have the largest weight. Spatial information may need to be considered as a part of color weights. In addition, the first step of the color transfer algorithm described in Section \ref{sec:cal} may result in cluster centers of the input image that do not have enough movement, depending on how colors are spread in the CIELAB space and where those target Pantone colors are.

%\begin{figure}[]
%\centering
%    \includegraphics[width = 84mm]{fig15.jpg}
%\caption{Limitation of clustering algorithm. In (b), black, gray and white are the dominant, subordinate and accent colors, respectively.}
%\label{fig:limit}
%\end{figure}

In the future, different color emotion models can be used in this framework, such as quantitative color emotion models.
In addition, in this paper we only used three color combinations, while adaptive number of colors may be used.

\vspace{-5mm}

%\begin{acknowledgements}
%\end{acknowledgements}

\bibliographystyle{spphys}       % APS-like style for physics
\bibliography{emotion-transfer}  % name your BibTeX data base

\end{document}